\documentclass[10pt,twocolumn,letterpaper]{article}

\usepackage{cvpr}
\usepackage{times}
\usepackage{epsfig}
\usepackage{graphicx}
\usepackage{amsmath}
\usepackage{amssymb}

\usepackage[breaklinks=true,bookmarks=false]{hyperref}

\cvprfinalcopy 

\def\cvprPaperID{****} 

\begin{document}

\title{Self-Supervised Deep Visual Odometry with Online Adaptation}

\author{Shunkai Li \qquad
	Xin Wang \qquad
	Yingdian Cao \qquad
	Fei Xue \qquad
	Zike Yan\qquad
	Hongbin Zha\\
	Key Laboratory of Machine Perception (MOE), School of EECS, Peking University\\
	PKU-SenseTime Machine Vision Joint Lab\\
	{\tt\small \{lishunkai, xinwang\_cis, yingdianc, feixue, zike.yan\}@pku.edu.cn \quad zha@cis.pku.edu.cn} 
}

\maketitle

\begin{abstract}
	Self-supervised VO methods have shown great success in jointly estimating camera pose and depth from videos. However, like most data-driven methods, existing VO networks suffer from a notable decrease in performance when confronted with scenes different from the training data, which makes them unsuitable for practical applications. In this paper, we propose an online meta-learning algorithm to enable VO networks to continuously adapt to new environments in a self-supervised manner. The proposed method utilizes convolutional long short-term memory (convLSTM) to aggregate rich spatial-temporal information in the past. The network is able to memorize and learn from its past experience for better estimation and fast adaptation to the current frame. When running VO in the open world, in order to deal with the changing environment, we propose an online feature alignment method by aligning feature distributions at different time. Our VO network is able to seamlessly adapt to different environments. Extensive experiments on unseen outdoor scenes, virtual to real world and outdoor to indoor environments demonstrate that our method consistently outperforms state-of-the-art self-supervised VO baselines considerably.
\end{abstract}

\section{Introduction}\label{introduction}
Simultaneous localization and mapping (SLAM) and visual odometry (VO) play a vital role for many real-world applications, such as autonomous driving, robotics and mixed reality. Classic SLAM/VO~\cite{DSO,LSD,svo,orb} methods perform well in regular scenes but fail in challenging conditions (\eg dynamic objects, occlusions, textureless regions) due to their reliance on low-level features. Since deep learning is able to extract high-level features and infer in an end-to-end fashion, learning-based VO~\cite{savo,beyond,GeoNet,SfMLearner} methods have been proposed in recent years to alleviate the limitation of classic hand-engineered algorithms.

\begin{figure}
	\begin{center}
		\includegraphics[width=1\linewidth]{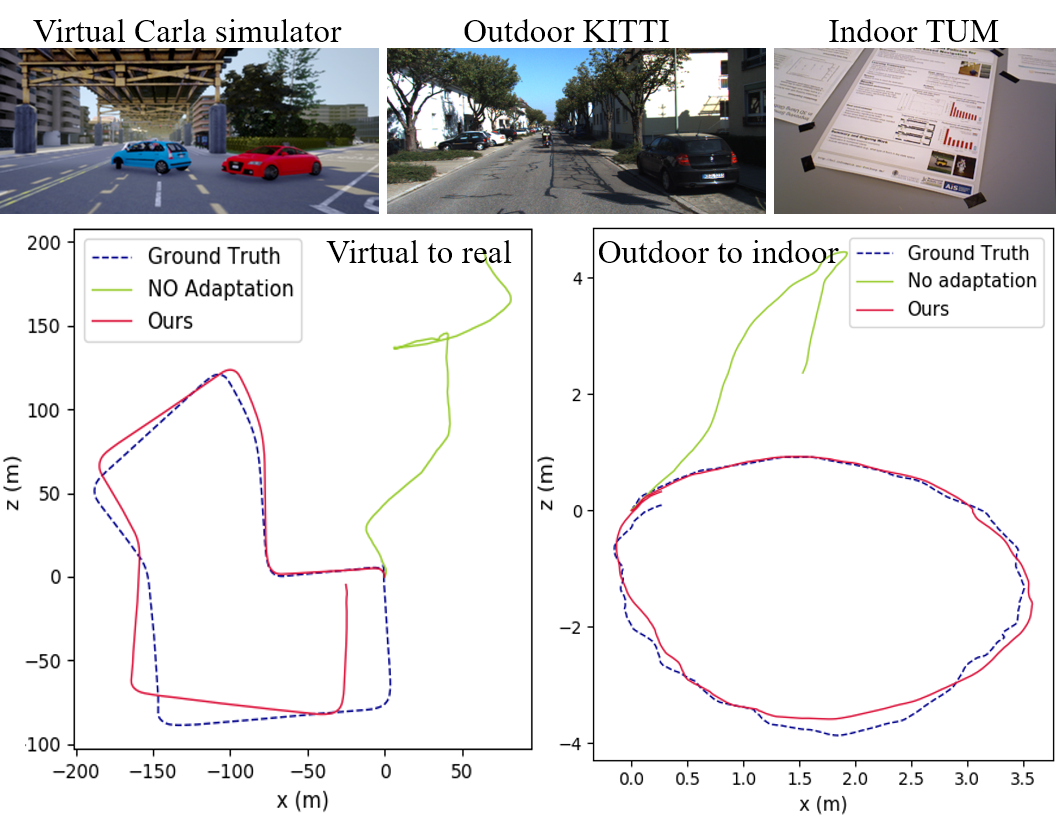}
	\end{center}
	\caption{We demonstrate the domain shift problem for self-supervised VO. Previous methods fail to generalize when the test data are different from the training data. In contrast, our method performs well when tested on changing environments, which demonstrates the advantage of fast online adaptation}
	\label{wo_online}
\end{figure}

However, learning-based VO suffers from a notable decrease in accuracy when confronted with scenes different from the training dataset~\cite{googleonline,realtime} (Fig.~\ref{wo_online}). When applied a pre-trained VO network to the \textit{open world}, the inability to generalize itself to new scenes presents a serious problem for its practical applications. This requires the VO network to continuously adapt to the new environment.

In contrast to fine-tuning a pre-trained network with ground truth data on the target domain~\cite{realtime}, it is unlikely to collect enough data in advance when running VO in the open world. This requires the network to adapt itself in \textit{real-time} to changing environments. In this online learning setting, there is no explicit distinction between training and testing phases --- \textbf{we learn as we perform}. This is much different from conventional learning methods where a pre-trained model is fixed during inference.

During online adaptation, the VO network can only learn from the current data instead of the entire training data with batch training and multiple epoches~\cite{slow}. The learning objective is to find an optimal model that is well adapted to the current data. However, because of the limited temporal perceptive field~\cite{forget2}, the current optimal model may not be well suited for subsequent frames. This makes the optimal parameters oscillate with time, leading to slow convergence during online adaptation~\cite{forget3,slow,forget1}.

In order to address these issues, we propose an online meta-learning scheme for self-supervised VO that achieves online adaptation. The proposed method motivates the network to perform consistently well at different time by incorporating online adaptation process into the learning objective. Besides, the past experience can be used to accelerate the adaptation to a new environment. Therefore, instead of learning only from the current data, we employ convolutional long short-term memory (convLSTM) to aggregate rich spatial-temporal information in the video that enables the network to use past experience for better estimation and also adapt quickly to the current frame. In order to achieve fast adaptation in changing environments, we propose a feature alignment method to align non-stationary feature distributions at different time. The proposed network automatically adapts to changing environments without ground truth data collected in advance for external supervision. Our contributions can be summarized as follows:
\begin{itemize}
	\item We propose an online meta-learning algorithm for VO to continuously adapt to unseen environments in a self-supervised manner.
	\item The VO network utilizes past experience incorporated by convLSTM to achieve better estimation and adapt quickly to the current frame.
	\item We propose a feature alignment method to deal with the changing data distributions in the open world.
\end{itemize}

Our VO network achieves 32 FPS on a Geforce 1080Ti GPU with online refinement, making it adapt in real-time for practical applications. We evaluate our algorithm across different domains, including outdoor, indoor and synthetic environments, which consistently outperforms state-of-the-art self-supervised VO baselines.

\section{Related works}
{\bf Learning-based VO} has been widely studied in recent years with the advent of deep learning and many methods with promising results have been proposed. Inspired by the framework of parallel tracking and mapping in classic SLAM/VO, DeepTAM~\cite{deeptam} utilizes two networks for pose and depth estimation simultaneously. DeepVO~\cite{deepvo} uses recurrent neural network (RNN) to leverage sequential correlations to estimate poses recurrently. However, these methods require ground truth which is expensive or impractical to obtain. To avoid the need of annotated data, self-supervised VO has been recently developed. SfMLearner~\cite{SfMLearner} utilizes the 3D geometric constraint of pose and depth to learn by minimizing photometric loss. Yin \etal~\cite{GeoNet} and Ranjan \etal~\cite{competitive} extend this idea to joint estimation of pose, depth and optical flow to handle non-rigid cases which are against static-scene assumption. These methods focus on mimicking local structure from motion (SfM) with image pairs, but fail to exploit spatial-temporal correlations over long sequence. SAVO~\cite{savo} formulates VO as a sequential generative task and utilizes RNN to reduce scale drift significantly. In this paper, we adopt the same idea as SfMLearner~\cite{SfMLearner} and SAVO~\cite{savo}.

{\bf Online adaptation} Most machine learning models suffer from a significant reduce in performance when the test data are different from the training set. An effective solution to alleviate this domain shift issue is online learning~\cite{lifelong}, where data are processed sequentially and data distribution changes continuously. Previous methods use online gradient update~\cite{onlineSGD} and probabilistic filtering~\cite{stream}. Recently, domain adaptation has been widely studied in computer vision. Long \etal~\cite{domainshift} propose Maximum Mean Discrepancy loss to reduce the domain shift. Several works~\cite{GANadapt2,GANadapt} utilize Generative Adversarial Networks (GAN) to directly transfer images in the target domain to the source domain (\eg day to night or winter to summer). Inspired by~\cite{GANadapt2, feature}, we propose a feature alignment method for online adaptation.

\begin{figure*}
	\begin{center}
		\includegraphics[width=0.96\linewidth]{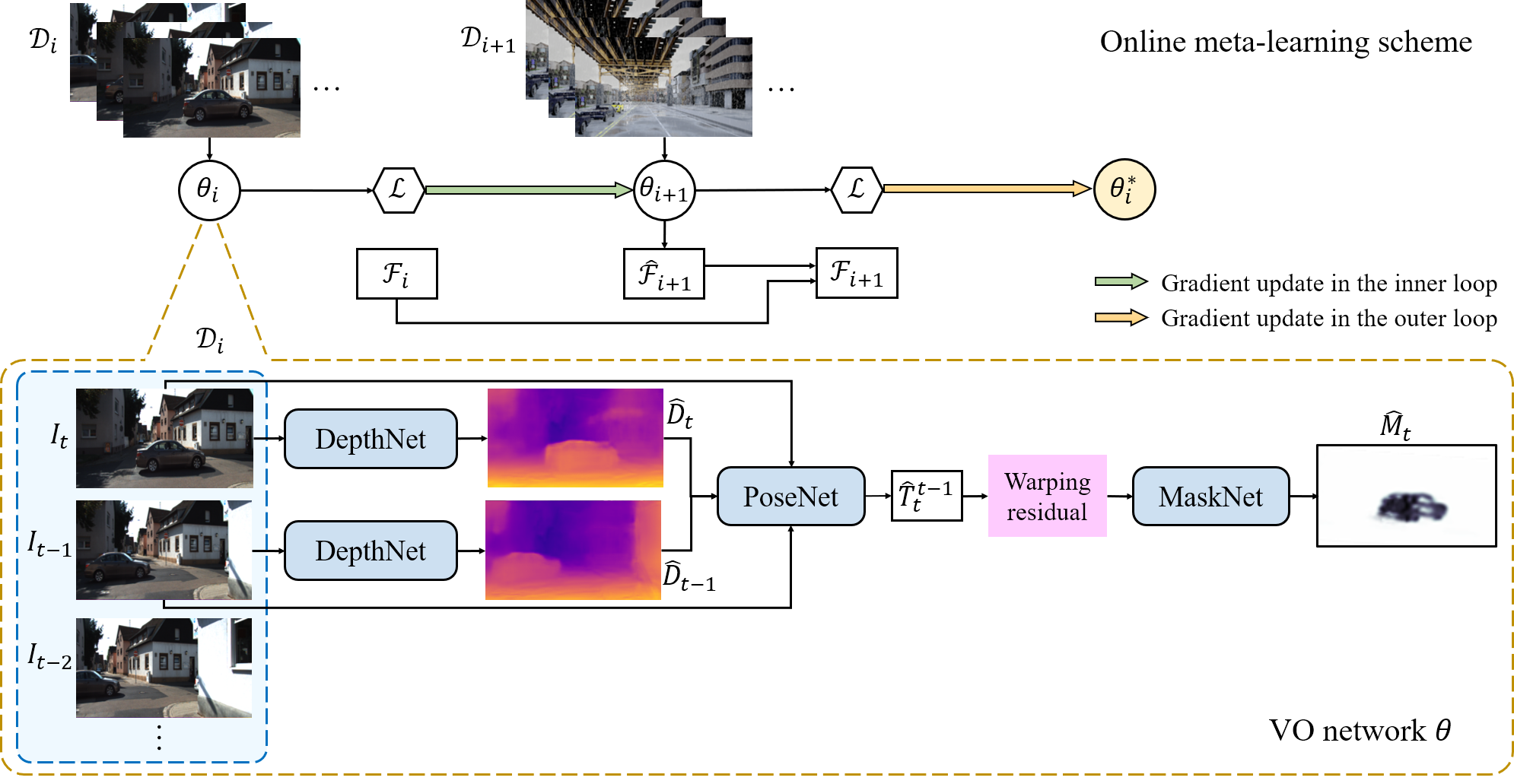}
	\end{center}
	\caption{The framework of our method. The VO network estimates pose $\hat{T}^{t-1}_{t}$, depth $\hat{D}_{t}, \hat{D}_{t-1}$ and mask $\hat{M}_{t}$ from image sequences $\mathcal{D}_{i}$. At each iteration $i$, the network parameters $\theta_{i}$ are updated according to the loss $\mathcal{L}$ and performs inference for $\mathcal{D}_{i+1}$ at next time. The network learns to find a set of weights $\theta^{*}_{i}$ that perform well both for $\mathcal{D}_{i}$ and $\mathcal{D}_{i+1}$. During online learning, spatial-temporal information is aggregated by convLSTM and feature alignment is adopted to align feature distributions $\hat{\mathcal{F}_{i}}, \hat{\mathcal{F}}_{i+1}$ at different time for fast adaptation}
	\label{framework}
\end{figure*}

{\bf Meta-learning}, or learning to learn, is a continued interest in machine learning. It exploits inherent structures in data to learn more effective learning rules for fast domain adaptation~\cite{meta2,meta}. A popular approach is to train a meta-learner that learns how to update the network~\cite{meta3,MAML}. Finn \etal~\cite{MAML,meta8} proposed Model Agnostic Meta-Learning (MAML) that constrains the learning rule for the model and uses stochastic gradient descent to quickly adapt networks to new tasks. This simple yet effective formulation has been widely used to adapt deep networks to unseen environments~\cite{meta7,meta5,meta4,meta6,meta9}. Our proposed method is most relevant to MAML, which extends it to the self-supervised, online learning setting.

\section{Problem setup}

\subsection{Self-supervised VO}\label{ssVO}
Our self-supervised VO follows the similar idea of SfMLearner~\cite{SfMLearner} and SAVO~\cite{savo} (shown in Fig.~\ref{framework}). The DepthNet predicts depth $\hat{D}_{t}$ of the current frame $I_{t}$. The PoseNet takes stacked monocular images $I_{t-1}, I_{t}$ and $\hat{D}_{t-1}, \hat{D}_{t}$ to regress relative pose $\hat{T}^{t-1}_{t}$. Then view synthesis is applied to reconstruct $\hat{I}_{t}$ by differentiable image warping:
\begin{equation}
p_{t-1} \sim K\hat{T}^{t-1}_{t}\hat{D}_{t}(p_{t})K^{-1}p_{t},
\label{warping}
\end{equation}
where $p_{t-1}, p_{t}$ are the homogeneous coordinates of a pixel in $I_{t-1}$ and $I_{t}$, respectively. $K$ denotes camera intrinsics. The MaskNet predicts a per-pixel mask $\hat{M}_{t}$~\cite{SfMLearner} according to the warping residuals $\|\hat{I}_{t}-I_{t}\|_{1}$.

\subsection{Online adaptation}\label{onlineadaptation}

As shown in Fig.~\ref{wo_online}, the performance of VO networks is fundamentally limited by their generalization ability when confronted with scenes different from the training data. The reason is they are designed under a \textbf{closed world} assumption: the training data $\mathcal{D}^{train}$ and test data $\mathcal{D}^{test}$ are i.i.d. sampled from a common dataset with fixed distribution. However, when running a pre-trained VO network in the \textbf{open world}, images are continuously collected in changing scenes. In this sense, the training and test data no longer share similar visual appearances, and the data at the current view may be different from previous views. This requires the network to online adapt to changing environments.

Given a model $\theta$ pretrained on $\mathcal{D}^{train}$, a naive approach for online learning is to update parameters $\theta$ by computing loss $\mathcal{L}$ on the current data $\mathcal{D}_{i}$:
\begin{equation}
\theta_{i+1} = \theta_{i} - \alpha \nabla_{\theta_{i}}\mathcal{L}(\theta_{i}, \mathcal{D}_{i}),
\label{update}
\end{equation}
where $\theta_{0}=\theta$ and $\alpha$ is the learning rate. Despite its simplicity, this approach has several drawbacks. The temporal perceptive field of the learning objective $\mathcal{L}(\theta_{i}, \mathcal{D}_{i})$ is 1, which means it accounts only for the current input $\mathcal{D}_{i}$ and has no correlation with previous data. The optimal solution for current $\mathcal{D}_{i}$ is likely to be unsuitable for subsequent inputs. Therefore, the gradients $\nabla_{\theta_{i}}\mathcal{L}(\theta_{i}, \mathcal{D}_{i})$ at  different iterations are stochastic without consistency~\cite{forget3,forget2}. This leads to slow convergence and may introduce negative bias in the learning procedure.

\section{Method}
In order to address these issues, we propose to exploit \textit{correlations} of different time for fast online adaptation. Our framework is illustrated in Fig.~\ref{framework}. The VO network $\theta_{i}$ takes $N$ consecutive frames in the sliding window $\mathcal{D}_{i}$ to estimate pose and depth in a self-supervised manner (Sec.~\ref{ssVO}). Then it is updated according to the loss $\mathcal{L}$ and infers for frames $\mathcal{D}_{i+1}$ at the next time. The network learns to find a set of weights $\theta^{*}_{i}$ to perform well both for $\mathcal{D}_{i}$ and $\mathcal{D}_{i+1}$ (Sec.~\ref{ssoml}). During online learning, spatial-temporal information is incorporated by convLSTM (Sec.~\ref{sta}) and feature alignment is adopted (Sec.~\ref{fa}) for fast adaptation.

\subsection{Self-supervised online meta-learning}\label{ssoml}
In contrast to $\mathcal{L}(\theta_{i}, \mathcal{D}_{i})$, we extend the online learning objective to $\mathcal{L}(\theta_{i+1}, \mathcal{D}_{i+1})$, which can be written as:
\begin{equation}
\min\limits_{\theta_{i}}\mathcal{L}(\theta_{i}-\alpha\nabla_{\theta_{i}}\mathcal{L}(\theta_{i}, \mathcal{D}_{i}), \mathcal{D}_{i+1}).
\label{our_objective}
\end{equation}
Different from naive online learning, the temporal perceptive field of Eq.~\ref{our_objective} becomes 2. It optimizes the performance on $\mathcal{D}_{i+1}$ after adapting to the task on $\mathcal{D}_{i}$. The insight is instead of minimizing the \textit{training} error $\mathcal{L}(\theta_{i}, \mathcal{D}_{i})$ on the current iteration $i$, we try to minimize the \textit{test} error on the next iteration. Our formulation directly incorporates online adaptation into the learning objective, which motivates the network to learn $\theta_{i}$ at $i$ to perform better at next time $i+1$.

Our objective of learning to adapt is similar in spirit to that of Model Agnostic Meta Learning (MAML)~\cite{MAML}:
\begin{equation}
\min\limits_{\theta}\sum_{\tau\in\mathcal{T}}\mathcal{L}(\theta-\alpha\nabla_{\theta}\mathcal{L}(\theta, \mathcal{D}_{\tau}^{train}), \mathcal{D}_{\tau}^{val}),
\label{MAML_objective}
\end{equation}
which aims to minimize the evaluation (adaptation) error on the validation set instead of minimizing the training error on the training set. $\tau$ denotes tasks sampled from the task set $\mathcal{T}$. More details of MAML can be found in~\cite{MAML}.

As a nested optimization problem, our objective function is optimized via a two-stage gradient descent. At each iteration $i$, we take $N$ consecutive frames in the sliding window as a mini-dataset $\mathcal{D}_{i}$ (shown within the blue area in Fig.~\ref{framework}):
\begin{equation}
\mathcal{D}_{i}=\{I_{t}, I_{t-1}, I_{t-2},\dots,I_{t-N+1}\}.
\end{equation}
In the inner loop of Eq.~\ref{our_objective}, we evaluate the performance of VO in $D_{i}$ by self-supervised loss $\mathcal{L}$ and update parameters $\theta_{i}$ according to Eq.~\ref{update}. Then, in the outer loop, we evaluate the performance of the updated model $\theta_{i+1}$ on subsequent frames $\mathcal{D}_{i+1}$. We mimic this continuous adaptation process on both training and online test phases. During training, we minimize the sum of losses by Eq.~\ref{our_objective} across all sequences in the training dataset, which motivates the network to learn base weights $\theta$ that enables fast online adaptation.

In order to provide more intuition on what it learns and the reason for fast adaptation, we take Taylor expansion on our training objective:
\begin{equation}
\begin{aligned}
&\min\limits_{\theta_{i}}\mathcal{L}(\theta_{i}-\alpha\nabla_{\theta_{i}}\mathcal{L}(\theta_{i}, \mathcal{D}_{i}), \mathcal{D}_{i+1}) \\
&\approx\min\limits_{\theta_{i}}\mathcal{L}(\theta_{i}, \mathcal{D}_{i+1})-\alpha\nabla_{\theta_{i}}\mathcal{L}(\theta_{i}, \mathcal{D}_{i})\cdot\nabla_{\theta_{i}}\mathcal{L}(\theta_{i}, \mathcal{D}_{i+1}) \\
&\quad+H_{\theta_{i}}\cdot[\alpha\nabla_{\theta_{i}}\mathcal{L}(\theta_{i}, \mathcal{D}_{i})]^{2}+\dots \\
&\approx\min\limits_{\theta_{i}}\mathcal{L}(\theta_{i}, \mathcal{D}_{i+1})-\alpha\left\langle\nabla_{\theta_{i}}\mathcal{L}(\theta_{i}, \mathcal{D}_{i}),\nabla_{\theta_{i}}\mathcal{L}(\theta_{i}, \mathcal{D}_{i+1})\right \rangle, \\
\end{aligned}
\label{Taylor}
\end{equation}
where $H_{\theta_{i}}$ denotes Hessian matrix and $\left\langle\cdot,\cdot\right\rangle$ denotes inner product. Since most neural networks use ReLU activations, the networks are locally linear, thus the second order derivative equals 0 in most cases~\cite{relu2nd}. Therefore, $H_{\theta_{t}}\approx0$ and higher order terms are also omitted.

As shown in Eq.~\ref{Taylor}, the network learns to minimize the prediction error $\mathcal{L}(\theta_{i}, \mathcal{D}_{i+1})$ with $\theta_{i}$ while maximizing the similarity between the gradients at $\mathcal{D}_{i}$ and $\mathcal{D}_{i+1}$. Since the camera is continuously moving, the scenes $\mathcal{D}_{i},\mathcal{D}_{i+1}$ may vary from different time. Naive online learning treats different scenes \textit{independently} by fitting only the current scene but ignores the way to perform VO in different scenes are similar. As gradient indicates the direction to update the network, this leads to inconsistent gradients at $i, i+1$ and slow convergence. In contrast, the second term enforces consistent gradient directions by \textit{aligning} gradient for $\mathcal{D}_{i+1}$ with previous information, indicating that we are training the network $\theta_{i}$ at $i$ to perform consistently well for both $i$ and $i+1$. This meta-learning scheme alleviates stochastic gradient problem in online learning. Eq.~\ref{Taylor} describes the dynamics of sequential learning in non-stationary scenes. The network learns to adjust at current state by $\mathcal{L}(\theta_{i}, \mathcal{D}_{i})$ to better perform at next time. Consequently, the learned $\theta$ is less sensitive to the non-stationary data distributions of sequential inputs, enabling fast adaptation to unseen environments.

\subsection{Spatial-temporal aggregation}\label{sta}
As stated in Sec.~\ref{introduction}, online learning suffers from slow convergence due to the inherent limitation of temporal perceptive field. In order to make online updating more effective, we let the network perform current estimation based on previous information. Besides, predicting pose from only image pairs is prone to error accumulation. This trajectory drift problem can be mitigated by exploiting spatial-temporal correlations over long sequence~\cite{savo,beyond}.

\begin{figure}[!h]
	\begin{center}
		\includegraphics[width=1\linewidth]{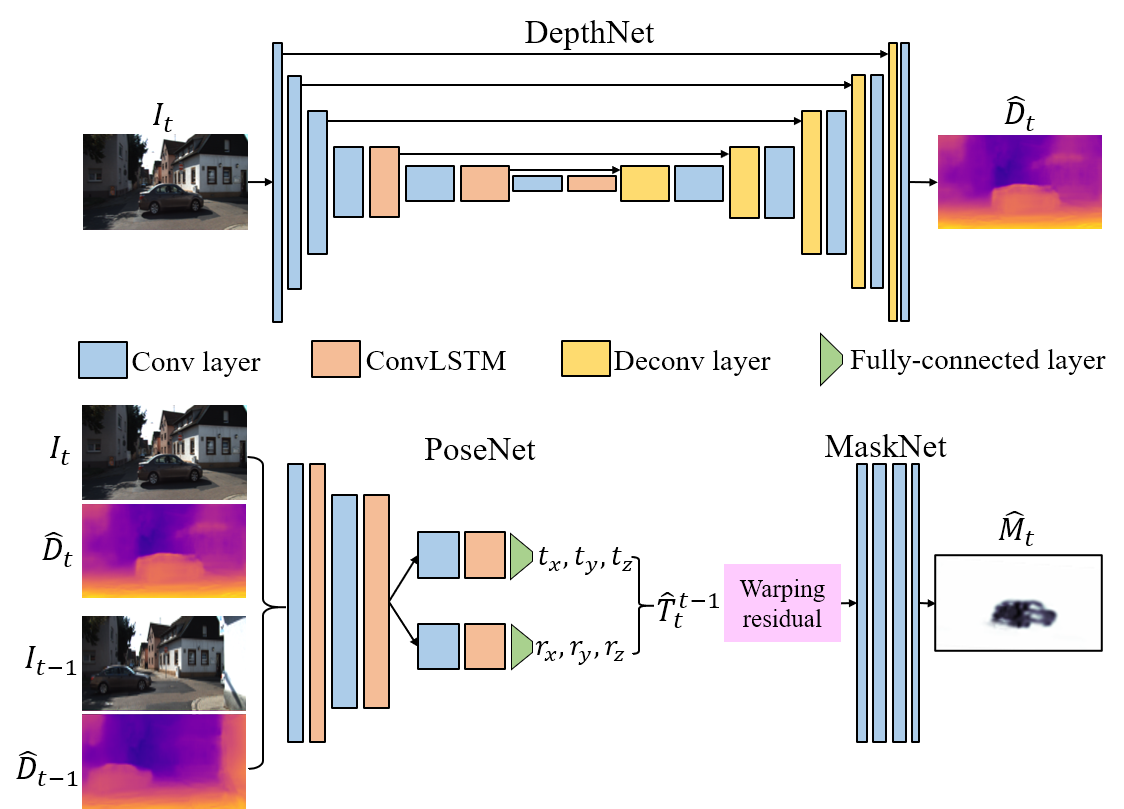}
	\end{center}
	\caption{Network architecture of DepthNet, PoseNet and MaskNet in self-supervised VO framework. The height of each block represents the size of its feature maps}
	\label{convLSTM}
\end{figure}

In this paper, we use convolutional LSTM (convLSTM) to achieve fast adaptation and reduce accumulated error. As shown in Fig.~\ref{convLSTM}, we embed recurrent units into the encoder of DepthNet and PoseNet to allow the convolutional network to leverage not only spatial but also temporal information for depth and pose estimation. The length $N$ of convLSTM is the number of frames in $\mathcal{D}_{i}$. ConvLSTM acts as the memory of the network. As new frames are processed, the network is able to memorize and learn from its past experience, so as to update parameters to quickly adapt to unseen environments. This approach not only enforces correlations among different time steps, but also learns the temporally dynamic nature of the moving camera from video inputs.

\subsection{Feature alignment}\label{fa}
One basic assumption of conventional machine learning is that the training and test data are independently and identically (i.i.d.) drawn from the same distribution. However, this assumption does not hold when running VO in the open world, since the test data (target domain) are usually different from the training data (source domain). Besides, as the camera is continuously moving in the changing environment, the captured scenes $\mathcal{D}_{i}$ also vary in time. As highlighted in~\cite{feature,feature2}, aligning feature distributions of two domains will improve performance in domain adaptation.

Inspired by~\cite{feature}, we extend this domain adaptation method to the online learning setting by aligning feature distributions in different time. When training on the source domain, we collect the statistics of features $f_{j}\in\{f_{1},...,f_{n}\}$ in a feature map tensor by Layer Normalization (LN)~\cite{LN}:
\begin{equation}
\begin{aligned}
\mathcal{F}_{s}&=(\mu_{s},\sigma^{2}_{s}), \\
\mu_{s}=\frac{1}{n}\sum_{j=1}^{n}f_{j},& \quad \sigma^{2}_{s}=\frac{1}{n}\sum_{j=1}^{n}(f_{j}-\mu_{s})^2, \\
n=&H\times W\times C, \\
\end{aligned}
\label{G_s}
\end{equation}
where $H, W, C$ are the height, width and channels of each feature map. When adapted to the target domain, we initialize feature statistics at $i=0$:
\begin{equation}
\mathcal{F}_{0}=\mathcal{F}_{s}.
\label{G_0}
\end{equation}
Then at each iteration $i$, feature statistics $\hat{\mathcal{F}}_{i}=(\hat{\mu_{i}}, \hat{\sigma^{2}_{i}})$ are computed by Eq.~\ref{G_s}. Given previous statistics $\mathcal{F}_{i-1}=(\mu_{i-1}, \sigma^{2}_{i-1})$, feature distribution at $i$ is aligned by:
\begin{equation}
\begin{aligned}
&\mu_{i}=(1-\beta)\mu_{i-1}+\beta\hat{\mu}_{i}, \\
&\sigma^{2}_{i}=(1-\beta)\sigma^{2}_{i-1}+\beta\hat{\sigma}^{2}_{i}, \\
\end{aligned}
\label{G_i}
\end{equation}
where $\beta$ is a hyperparameter. After feature alignment, the features $f_{j}\in\{f_{1},...,f_{n}\}$ are normalized to~\cite{LN}:
\begin{equation}
\hat{f}_{j}=\gamma\frac{f_{j}-\mu_{i}}{\sqrt{\sigma^{2}_{i}+\epsilon}}+\delta,
\end{equation}
where $\epsilon$ is a small constant for numerical stability. $\gamma$ and $\delta$ are the learnable scale and shift in normalization layers~\cite{LN}.

The insight of this approach is to enforce correlation of non-stationary feature distributions in changing environments. Learning algorithms perform well when feature distribution of the test data is the same as the training data. When changed to a new environment, despite the extracted features are different, we deem that feature distributions of two domains should be the same (Eq.~\ref{G_0}). Despite the view is changing when running VO in an open world, $\mathcal{D}_{i}$ and $\mathcal{D}_{i+1}$ are observed continuously in time, thus their feature distributions should be similar (Eq.~\ref{G_i}). This feature normalization and alignment approach acts as \textit{regularization} that simplifies the learning process, which makes the learned weights $\theta$ consistent for non-stationary environments.

\subsection{Loss functions}
Our self-supervised loss $\mathcal{L}$ is the same as most previous methods. It consists of:

{\bf Appearance loss} We measure the reconstructed image $\hat{I}$ by photometric loss and structural similarity metric (SSIM):
\begin{equation}
\begin{aligned}
\mathcal{L}_{a}&=\lambda_{m}\mathcal{L}_{m}(\hat{M})+(1-\alpha_{s})\frac{1}{N}\sum\hat{M}\|\hat{I}-I\|_{1} \\
&+\frac{1}{N}\sum_{x,y}\alpha_{s}\frac{1-\text{SSIM}(\hat{I}(x,y), I(x,y))}{2}. \\
\end{aligned}
\end{equation}
The regularization term $\mathcal{L}_{m}(\hat{M})$ prevents the learned mask $\hat{M}$ converges to a trivial solution~\cite{SfMLearner}. The filter size of SSIM is set 5$\times$5 and $\alpha_{s}$ is set 0.85.

{\bf Depth regularization} We introduce an edge-aware loss to enforce discontinuity and local smoothness in depth:
\begin{equation}
\begin{aligned}
\mathcal{L}_{r}=&\frac{1}{N}\sum_{x,y}\|\nabla_{x}\hat{D}(x,y)\|e^{-\|\nabla_{x}I(x,y)\|}+ \\&\|\nabla_{y}\hat{D}(x,y)\|e^{-\|\nabla_{y}I(x,y)\|}.
\end{aligned}
\end{equation}

Thus the self-supervised loss $\mathcal{L}$ is:
\begin{equation}
\mathcal{L}=\lambda_{a}\mathcal{L}_{a}+\lambda_{r}\mathcal{L}_{r}.
\end{equation}

\section{Experiments}
\subsection{Implementation details}\label{imp}
The architecture of our network is shown in Fig.~\ref{convLSTM}. The DepthNet uses a U-shaped architecture similar to~\cite{SfMLearner}. The PoseNet is splited into 2 parts followed by fully-connected layers to regress Euler angles and translations of 6-DoF pose, respectively. The length of convLSTM $N$ is set 9. Layer Normalization and ReLUs are adopted in each layer except for the output layers. Detailed network architecture can be found in the supplementary materials.

Our model is implemented by PyTorch~\cite{pytorch} on a single NVIDIA GTX 1080Ti GPU. All sub-networks are jointly trained in a self-supervised manner. Images are resized to 128$\times$416 during both training and online adaptation. The Adam~\cite{adam} optimizer with $\beta_{1}=0.9$, $\beta_{2}=0.99$ is used and the weight decay is set $4\times10^{-4}$. Weighting factors $\lambda_{m}, \lambda_{a}, \lambda_{r}$ are set 0.01, 1 and 0.5, respectively. The feature alignment parameter $\beta$ is set 0.5. The batch size is 4 for training and 1 for online adaptation. The learning objective (Eq.~\ref{our_objective}) is used for both training and online adaptation. We pre-train the network for 20,000 iterations. The learning rate $\alpha$ of the inner loop and outer loop are both initialized to $10^{-4}$ and reduced by half for every 5,000 iterations.

\subsection{Outdoor KITTI}\label{kitti}
First, we test our method on KITTI odometry~\cite{kitti} dataset. It contains 11 driving scenes with ground truth poses. We follow the same train/test split as~\cite{savo,GeoNet,SfMLearner} using sequences 00-08 for training and 09-10 for online test.

\begin{figure}
	\begin{center}
		\includegraphics[width=1\linewidth]{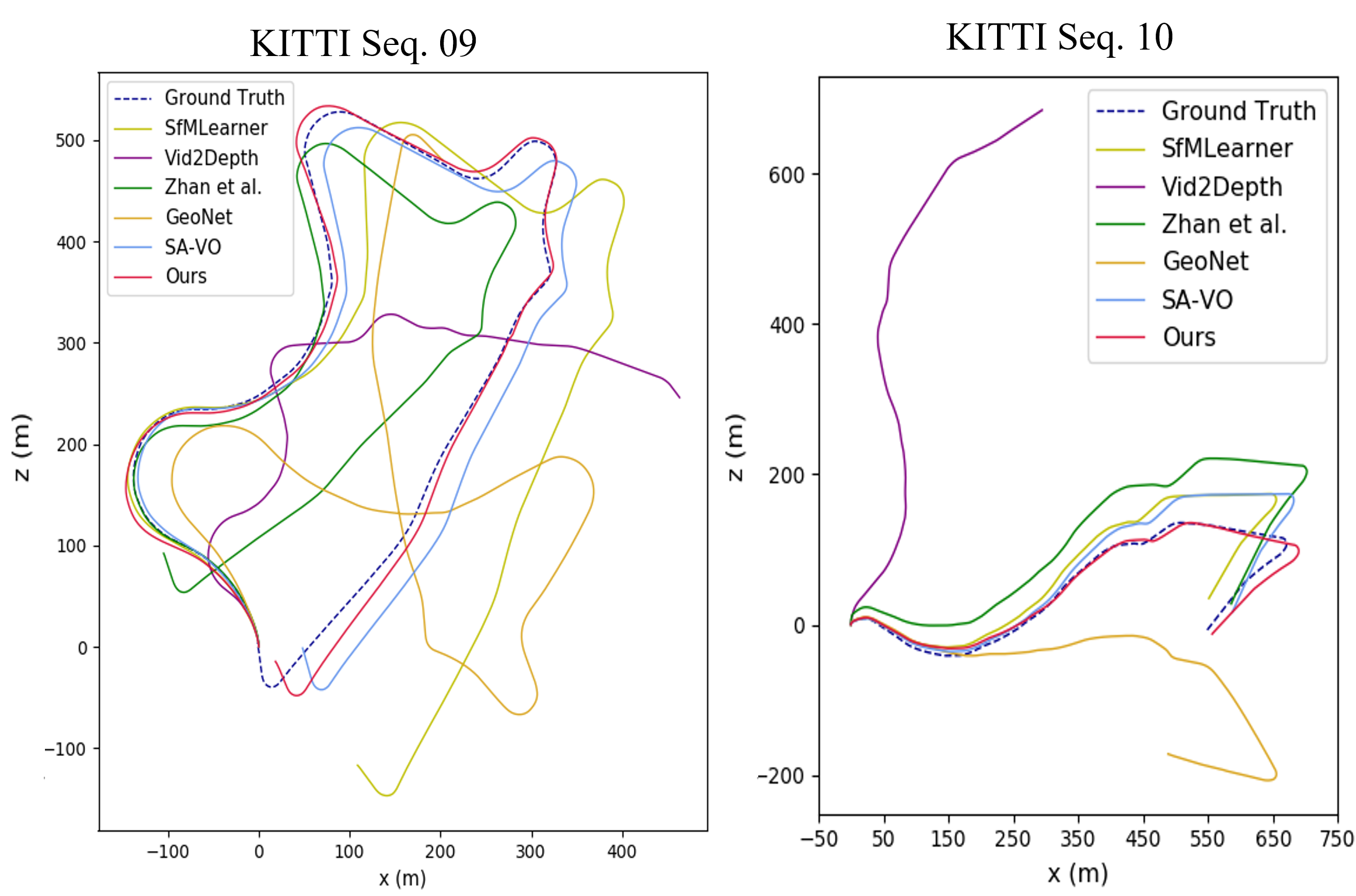}
	\end{center}
	\caption{Trajectories of different methods on KITTI dataset. Our method shows a better odometry estimation due to online updating}
	\label{KK}
\end{figure}

\begin{table}
	\small
	\begin{center}
		\begin{tabular}{lcccc}
			\hline
			\hline
			Method & \multicolumn{2}{c}{Seq. 09} & \multicolumn{2}{c}{Seq. 10} \\
			& $t_{err}$ & $r_{err}$ & $t_{err}$ & $r_{err}$ \\
			\hline
			SfMLearner~\cite{SfMLearner} & 11.15 & 3.72 & 5.98 & 3.40   \\
			Vid2Depth~\cite{vid2depth} & 44.52 & 12.11 & 21.45 & 12.50 \\
			Zhan \etal~\cite{deepvofeat} & 11.89 & 3.62 & 12.82 & 3.40 \\
			GeoNet~\cite{GeoNet} & 23.94 & 9.81 & 20.73 & 9.10  \\
			SAVO~\cite{savo} & 9.52 & 3.64 & 6.45 & 2.41 \\
			Ours & \textbf{5.89} & \textbf{3.34} & \textbf{4.79} & \textbf{0.83} \\
			\hline
			\hline
		\end{tabular}
	\end{center}
	\caption{Quantitative comparison of visual odometry results on
		KITTI dataset. $t_{err}$: average translational root mean square error (RMSE) drift (\%); $r_{err}$: average rotational RMSE drift (${}^{\circ}$/100m)}
	\label{K_table}
\end{table}

Instead of calculating absolute trajectory error (ATE) on image pairs in previous methods, we recover full trajectories and compute translation error $t_{err}$ by KITTI evaluation toolkit, rotation error $r_{err}$. We compare our method with several state-of-the-art self-supervised VO baselines: SfMLearner~\cite{SfMLearner}, GeoNet~\cite{GeoNet}, Zhan \etal~\cite{deepvofeat}, Vid2Depth~\cite{vid2depth} and SAVO~\cite{savo}. As stated in ~\cite{vid2depth}, a scaling factor is used to align trajectories with ground truth to solve the scale ambiguity problem in monocular VO. The estimated trajectories of sequences 09-10 are plotted in Fig.~\ref{KK} and quantitative evaluations are shown in Table~\ref{K_table}. Our method outperforms all the other baselines by a clear margin, the accumulated error is reduced by online adapation.

The comparison of the running speed with other VO methods can be found in Table~\ref{speed}. Since we are studying the online learning problem, the running time includes forward propagation, loss computing, back propagation and network updating. Our method achieves real-time online adaptation and outperforms state-of-the-art baselines considerably.

\begin{table}[h]
	\small
	\begin{center}
		\setlength{\tabcolsep}{1.2mm}{
		\begin{tabular}{l|ccccc}
			\hline
			Method & SfMLearner & GeoNet & Vid2Depth & SAVO & Ours\\
			\hline
			FPS & 24 & 21 & \textbf{37} & 17 & 32 \\
			\hline
		\end{tabular}}
	\end{center}
	\caption{Running speed of different VO methods.}
	\label{speed}
\end{table}

\subsection{Synthetic to real}\label{s2r}
Synthetic datasets (\eg virtual KITTI, Synthia and Carla) have been widely used for research since they provide ground truth labels and controllable environment settings. However, there's a large gap between the synthetic and real-world data. In order to test the domain adaptation ability, we use Carla simulator~\cite{carla} to collect synthetic images under different weather conditions in the virtual city for training, and use KITTI 00-10 for online testing.

\begin{figure*}
	\begin{center}
		\includegraphics[width=1\linewidth]{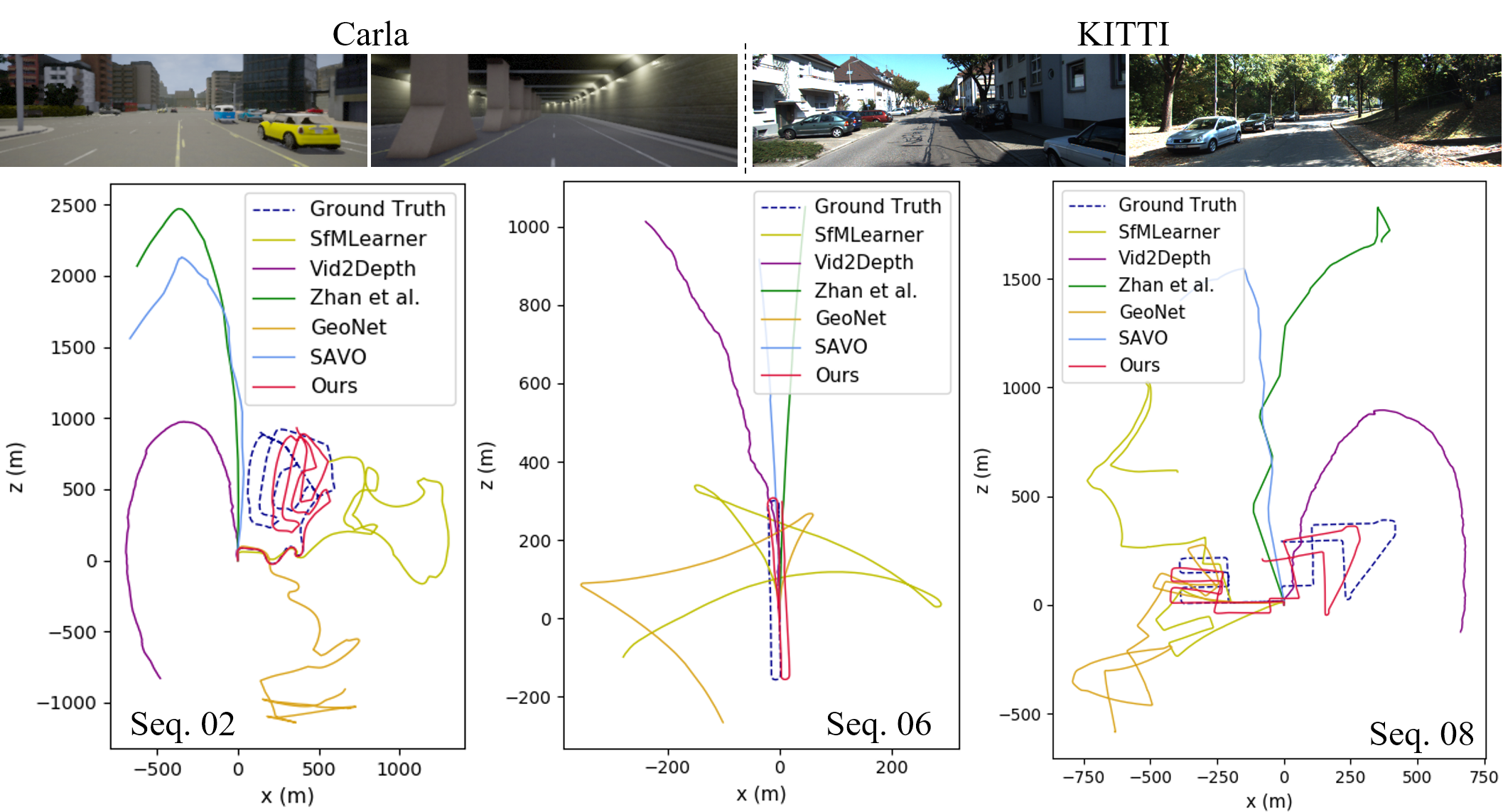}
	\end{center}
	\caption{Trajectories of different methods pretrained on Carla and test on KITTI dataset. Our method significantly outperforms all the other baselines when changed from virtual to the real-world data}
	\label{carla2kitti_figure}
\end{figure*}

It can be seen from Fig.~\ref{wo_online},~\ref{carla2kitti_figure} and Table~\ref{C2K_table} that previous methods all failed when shifted to real-world environments. This is probably because the features of virtual scenes are much different from the real world despite they are both collected in the driving scenario. In contrast, our method significantly outperforms previous arts, which is able to bridge the domain gap and quickly adapt to the real-world data.

\begin{table*}
	\footnotesize
	\small
	\begin{center}
		\begin{tabular}{cc|cccccccccccc}
			\hline
			\hline
			& & \multicolumn{2}{c}{SfMLearner~\cite{SfMLearner}}
			&   \multicolumn{2}{c}{Vid2Depth~\cite{vid2depth}}
			&   \multicolumn{2}{c}{Zhan \etal~\cite{deepvofeat}}
			&   \multicolumn{2}{c}{GeoNet~\cite{GeoNet}}
			&   \multicolumn{2}{c}{SAVO~\cite{savo}}
			&   \multicolumn{2}{c}{Ours} \\
			Seq & frames & $t_{err}$ & $r_{err}$ & $t_{err}$ & $r_{err}$
			& $t_{err}$ & $r_{err}$ & $t_{err}$ & $r_{err}$
			& $t_{err}$ & $r_{err}$ & $t_{err}$ & $r_{err}$\\
			\hline
			00 & 4541 & 61.55 & 27.13 & 61.69 & 28.41 & 63.30 & 28.24 & 44.08 & 14.89 & 60.10 & 28.43 & \textbf{14.21} & \textbf{5.93} \\
			01 & 1101 & 83.91 & 10.36 & 48.44 & 10.30 & 35.68 & 9.78 & 43.21 & 8.42 & 64.68 & 9.91 & \textbf{21.36} & \textbf{4.62} \\
			02 & 4661 & 71.48 & 27.80 & 70.56 & 25.72 & 84.63 & 24.67 & 73.59 & 12.53 & 69.15 & 24.78 & \textbf{16.21} & \textbf{2.60} \\
			03 & 801  & 49.51 & 36.81 & 41.92 & 27.31 & 50.05 & 16.44 & 43.36 & 14.56 & 66.34 & 16.45 & \textbf{18.41} & \textbf{0.89} \\
			04 & 271  & 23.80 & 10.52 & 39.34 & 3.42 & 12.08 & \textbf{1.56} & 17.91 & 9.95 & 25.28 & 1.84 & \textbf{9.08} & 4.41 \\
			05 & 2761 & 87.72 & 30.71 & 63.62 & 30.71 & 89.03 & 29.66 & 32.47 & 13.12 & 59.90 & 29.67 & \textbf{24.82} & \textbf{6.33} \\
			06 & 1101 & 59.53 & 12.70 & 84.33 & 32.75 & 93.66 & 30.91 & 40.28 & 16.68 & 63.18 & 31.04 & \textbf{9.77} & \textbf{3.58} \\
			07 & 1101 & 51.77 & 18.94 & 74.62 & 48.89 & 99.69 & 49.08 & 37.13 & 17.20 & 63.04 & 49.25 & \textbf{12.85} & \textbf{2.30} \\
			08 & 4701 & 86.51 & 28.13 & 70.20 & 28.14 & 87.57 & 28.13 & 33.41 & 11.45 & 62.45 & 27.11 & \textbf{27.10} & \textbf{7.81} \\
			09 & 1591 & 58.18 & 20.03 & 69.20 & 26.18 & 83.48 & 25.07 & 51.97 & 13.02 & 67.06 & 25.76 & \textbf{15.21} & \textbf{5.28} \\
			10 & 1201 & 45.33 & 16.91 & 49.10 & 23.96 & 53.70 & 22.93 & 46.63 & 13.80 & 58.52 & 23.02 & \textbf{25.63} & \textbf{7.69} \\
			\hline
			\hline
		\end{tabular}
	\end{center}
	\caption{Quantitative comparisons of different methods pretraining on synthetic data in Carla simulator and testing on KITTI}
	\label{C2K_table}
\end{table*}

\subsection{Outdoor KITTI to indoor TUM}\label{o2i}
\begin{figure*}
	\begin{center}
		\includegraphics[width=1\linewidth]{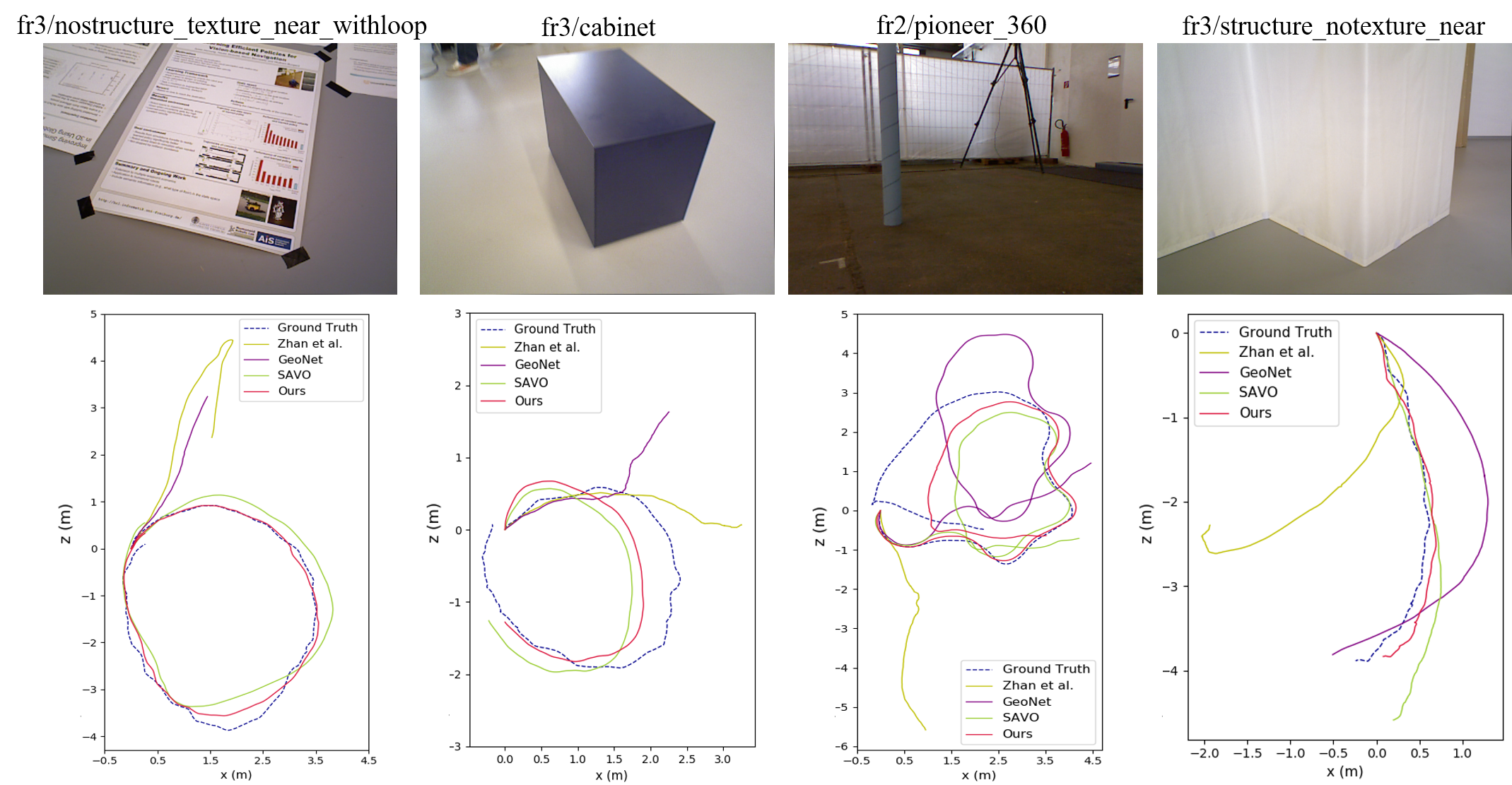}
	\end{center}
	\caption{Raw images (top) and trajectories (bottom) recovered by different methods on TUM-RGBD dataset}
	\label{KTUM}
\end{figure*}

In order to further evaluate the adaptability of our method, we test various baselines on TUM-RGBD~\cite{TUM} dataset. KITTI is captured by moving cars with planar motion, high quality images and sufficient disparity. Instead, TUM dataset is collected by handheld cameras in indoor scenes with much more complicated motion patterns, which is significantly different from KITTI. It includes various challenging conditions (Fig.~\ref{KTUM}) such as dynamic objects, non-texture scenes, abrupt motions and large occlusions.

We pretrain these methods on KITTI 00-08 and test on TUM dataset. Despite the ground truth depth is available, we only use monocular RGB images during test. It can be seen (Table~\ref{TUM_table} and Fig.~\ref{KTUM}) that our method consistently outperforms all the other baselines. Despite the large domain shift and significant difference in motion patterns (\ie large, planar motion vs small motion in 3 axes), our method can still recover trajectories well. On the contrary, GeoNet~\cite{GeoNet} and Zhan \etal~\cite{deepvofeat} tend to fail. Despite SAVO~\cite{savo} utilizes LSTM to alleviate accumulated error to some extent, our method performs better due to online adaptation.

\begin{table*}
	\begin{center}
		\begin{tabular}{l|ccc|cccc}
			\hline
			\hline
			Sequence & Structure & Texture & Abrupt motion & Zhan \etal~\cite{deepvofeat} & GeoNet~\cite{GeoNet} & SAVO~\cite{savo} & Ours \\
			\hline
			fr2/desk & \checkmark & \checkmark & - & 0.361 & 0.287 & 0.269 & \textbf{0.214} \\ 
			fr2/pioneer\_360  & \checkmark & \checkmark & \checkmark & 0.306 & 0.410 & 0.383 & \textbf{0.218} \\ 
			fr2/pioneer\_slam & \checkmark & \checkmark & \checkmark & 0.309 & 0.301 & 0.338 & \textbf{0.190} \\ 
			fr2/360\_kidnap & \checkmark & \checkmark & \checkmark & 0.367 & 0.325 & 0.311 & \textbf{0.298} \\ 
			fr3/cabinet 	  & \checkmark & - & - & 0.316 & 0.282 & 0.281 & \textbf{0.272} \\     
			fr3/long\_off\_hou\_valid & \checkmark & \checkmark & - & 0.327 & 0.316 & 0.297 & \textbf{0.237} \\ 
			fr3/nstr\_tex\_near\_loop & - & \checkmark & - & 0.340 & 0.277 & 0.440 & \textbf{0.255} \\ 
			fr3/str\_ntex\_far & \checkmark & - & - & 0.235 & 0.258 & 0.216 & \textbf{0.177} \\ 
			fr3/str\_ntex\_near & \checkmark & - & - & 0.217 & 0.198 & 0.204 & \textbf{0.128} \\ 
			\hline
			\hline
		\end{tabular}
	\end{center}
	\caption{Quantitative evaluation of different methods pretraining on KITTI and testing on TUM-RGBD dataset. We evaluate relative pose error (RPE) which is presented as translational RMSE in [m/s]}
	\label{TUM_table}
\end{table*}

\subsection{Ablation studies}\label{as}

In order to demonstrate the effectiveness of each component, we present ablation studies on various versions of our method on KITTI dataset (shown in Table~\ref{K_ablation_table}).

First, we evaluate the backbone of our method (the first row) which includes convLSTM and feature alignment but no meta-learning process during training and online test. It can be seen from Table~\ref{K_table} and Table~\ref{K_ablation_table} that, even without meta-learning and online adaptation, our network backbone still outperforms most pervious methods. The results indicate that convLSTM is able to reduce accumulated error and feature alignment improves the performance when confronted with unseen environments.

\begin{table}
	\small
	\begin{center}
		\setlength{\tabcolsep}{1.5mm}{
			\begin{tabular}{cccc|cccc}
				\hline
				\hline
				& & & & \multicolumn{2}{c}{Seq. 09} & \multicolumn{2}{c}{Seq. 10} \\
				Online & Pretrain & LSTM & FA & $t_{err}$ & $r_{err}$ & $t_{err}$ & $r_{err}$ \\

				\hline
				-			& Standard & \checkmark & \checkmark & 10.93 & 3.91 & 11.65 & 4.11 \\
				Naive    	& Standard & \checkmark & \checkmark & 10.22 & 5.33 & 8.24 & 3.22 \\
				\hline
				Meta		& Meta & - & - & 9.25 & 4.20 & 7.58 & 3.13 \\
				Meta		& Meta & \checkmark & - & 6.36 & 3.84 & 5.37 & 1.41 \\
				Meta		& Meta & - & \checkmark & 7.52 & 4.12 & 5.98 & 2.72  \\
				Meta		& Meta & \checkmark & \checkmark & \textbf{5.89} & \textbf{3.34} & \textbf{4.79} & \textbf{0.83} \\
				\hline
				\hline
		\end{tabular}}
	\end{center}
	\caption{Quantitative comparison of ablation study on KITTI dataset for various versions of our method. FA: feature alignment}
	\label{K_ablation_table}
\end{table}

Then we compare the efficiency of naive online learning (the second row) and meta-learning (the last row). It can be seen that, although naive online learning is able to reduce estimation error to some extent, it converges much slower than the meta-learning scheme, indicating that it takes much longer time to adapt the network to the new environment.

Finally, we study the effect of convLSTM and feature alignment during meta-learning (last four rows). Compared with baseline meta-learning scheme, convLSTM and feature alignment give the VO performance a further boost. Besides, convLSTM tends to perform better than feature alignment during online adaptation. One possible explaination is convLSTM incorporates spatial-temporal correlations and past experience over long sequence. It associates different states recurrently, making the gradient computation graph more intensively connected during back propagation. Meanwhile, convLSTM correlates the VO network at different time, enforcing to learn a set of weights $\theta$ that are consistent in the dynamic environment.

Besides, we study how the size of sliding window $N$ is influencing the VO performance. The change of $N$ has no much impact on the running speed (30-32 FPS), but as $N$ increases, the adaptation gets faster and better. When $N$ is greater than 15, the adaptation speed and accuracy becomes lower. Therefore, we set $N=15$ as the best choice.

\section{Conclusions}
In this paper, we propose an online meta-learning scheme for self-supervised VO to achieve fast online adaptation in the open world. We use convLSTM to aggregate spatial-temporal information in the past, enabling the network to use past experience for better estimation and fast adaptation to the current frame. Besides, we put forward a feature alignment method to deal with changing feature distributions in the unconstrained open world setting. Our network dynamically evolves in time to continuously adapt to changing environments on-the-fly. Extensive experiments on outdoor, virtual and indoor datasets demonstrate that our network with online adaptation ability outperforms state-of-the-art self-supervised VO methods.

\textbf{Acknowledgments}
This work is supported by the National Key Research and Development Program of China (2017YFB1002601) and National Natural Science Foundation of China (61632003, 61771026).

{\small
\bibliographystyle{ieee_fullname}
\bibliography{egbib}
}

\end{document}